\documentclass[11pt,a4paper]{article}
\usepackage{conll2018}
\usepackage{times}
\usepackage{latexsym}

\usepackage[utf8]{inputenc}
\usepackage{amssymb,amsfonts,amsmath}
\usepackage{graphicx}
\usepackage{multirow}
\usepackage{lipsum}
\usepackage{xcolor}
\usepackage{pgfplots,pgfplotstable}
\usepackage{pgfkeys}
\usepackage{float}
\usepackage{wrapfig}
\restylefloat{figure}
\usepackage{booktabs}
\usepackage[english]{babel}

\usepackage{caption}
\usepackage[flushleft]{threeparttable}
\usepackage{subcaption}
\usepackage{footnote}
\makesavenoteenv{tabular}
\makesavenoteenv{table}
\usepackage{scrextend}

\usepackage[linesnumbered,ruled]{algorithm2e}
\usepackage{setspace}
\usepackage{natbib}
\usepackage{array}
\usepackage{calc}
\usepackage{url}

\DeclareMathOperator*{\argmin}{arg\,min~}
\DeclareMathOperator*{\argmax}{arg\,max~}

\pgfplotsset{
	invoke before crossref tikzpicture={\tikzexternaldisable},
	invoke after crossref tikzpicture={\tikzexternalenable},
}

\usepgfplotslibrary{groupplots} 
\usetikzlibrary{matrix}
\usetikzlibrary{spy}
\usetikzlibrary{patterns}
\usepgfplotslibrary{external} 
\usepgfplotslibrary{units}
\usepgfplotslibrary{dateplot} 
\pgfplotsset{compat=newest}
\usetikzlibrary{patterns}

\definecolor{Turquoise}{RGB}{141, 211, 199}
\definecolor{darkgreen}{RGB}{0,100,0}

\newcommand*{\regularbox}[1]{\color{black}
	\setlength{\fboxsep}{-1\fboxrule}
	\fbox{\hspace{1.1pt}\textbf{\strut#1}\hspace{1.2pt}}
	\color{black}  }

\usepackage[nameinlink,capitalize]{cleveref}
\crefformat{appendix}{#2#1#3}

\newcommand{\x}{\mathbf{x}}
\newcommand{\y}{\mathbf{y}}
\newcommand{\z}{\mathbf{z}}

\newcommand{\s}{\mathbf{s}}

\newcommand{\E}{\mathbf{E}}

\aclfinalcopy 

\title{Active Learning for Interactive Neural \\ Machine Translation of Data Streams}

\author{\'{A}lvaro Peris \and Francisco Casacuberta \\
	Pattern Recognition and Human Language Technology Research Center \\ Universitat Politècnica de València, València, Spain\\
	\texttt{\{lvapeab, fcn\}@prhlt.upv.es}}

\date{}

\begin{document}
	\maketitle
	\begin{abstract}

		We study the application of active learning techniques to the translation of unbounded data streams via interactive neural machine translation. The main idea is to select, from an unbounded stream of source sentences, those worth to be supervised by a human agent. The user will interactively translate those samples. Once validated, these data is useful for adapting the neural machine translation model.
		
		We propose two novel methods for selecting the samples to be validated. We exploit the information from the attention mechanism of a neural machine translation system. Our experiments show that the inclusion of active learning techniques into this pipeline allows to reduce the effort required during the process, while increasing the quality of the translation system. Moreover, it enables to balance the human effort required for achieving a certain translation quality. Moreover, our neural system outperforms classical approaches by a large margin.

	\end{abstract}
	
	\section{Introduction}

The translation industry is a high-demand field.
Large amounts of data must be translated on a regular basis. 
Machine translation (MT) techniques greatly boost the productivity of the translation agencies \citep{Arenas08}. However, despite the recent advances achieved in this field, MT systems are still far to be perfect and make errors. 
The correction of such errors is usually done in a post-processing step, called post-editing. This requires a great effort, as it needs from expert human supervisors.

The requirements of the translation industry have increased in the last years. We live in a global world, in which large amounts of data must be periodically translated. This is the case of the European Parliament, whose proceedings must be regularly translated; or the Project Syndicate\footnote{\url{www.project-syndicate.org}} platform, which translates editorials from newspapers to several languages. In these scenarios, the sentences to be translated can be seen as unbounded streams of data \citep{Levenberg10}.

When dealing with such massive volumes of data, it is prohibitively expensive to manually revise all the translations. Therefore, it is mandatory to spare human effort, at the expense of some translation quality. Hence, when facing this situation, we have a twofold objective: on the one hand, we aim to obtain translations with the highest quality possible. On the other hand, we are constrained by the amount of human effort spent in the supervision and correction process of the translations proposed by an MT system.

The active learning (AL) framework is well-suited for these objectives. The application of AL techniques to MT involve to ask a human oracle to supervise a fraction of the incoming data \citep{Bloodgood10}. Once the human has revised these samples, they are used for improving the MT system, via incremental learning. Therefore, a key element of AL is the so-called sampling strategy, which determines the sentences that should be corrected by the human. 

Aiming to reduce the human effort required during post-editing, other alternative frameworks have been study. A successful one is the interactive-predictive machine translation (IMT) paradigm \citep{Foster97,Barrachina09}. In IMT, human and MT system jointly collaborate for obtaining high-quality translations, while reducing the human effort spent in this process.



%

In this work, we explore the application of NMT to the translation of unbounded data streams. We apply AL techniques for selecting the instances to be revised by a human oracle. The correction process is done by means of an interactive-predictive NMT (INMT) system, which aims to reduce the human effort of this process. The supervised samples will be used for the NMT system to incrementally improve its models.
To the best of our knowledge, this is the first work that introduces an INMT system into the scenario involving the translation of unbounded data. Our main contributions are:
\begin{itemize} 
\item We study the application of AL on an INMT framework when dealing with large data streams. We introduce two sampling strategies for obtaining the most useful samples to be supervised by the human. We compare these techniques with other classical, well-performing strategies. 
\item  We conduct extensive experiments, analyzing the different sampling strategies and studying the amount of effort required for obtaining a certain translation quality. 
\item  The results show that AL succeeds at improving the translation pipeline. The translation systems featuring AL have better quality and require less human effort in the IMT process than static systems. Moreover, the application of the AL framework allows to obtain a balance between translation quality and effort required for achieving such quality. This balance can be easily tuned, according to the needs of the users.
\item  We open-source our code\footnote{The source code can be found at:\\\href{http://github.com/lvapeab/nmt-keras/tree/interactive_NMT}{http://github.com/lvapeab/nmt-keras}.} and use publicly-available corpora, fostering further research on this area.
\end{itemize}

	\section{Related work}
\label{sec:related-work}

The translation of large data streams is a problem that has been thoroughly studied. Most works aim to continuously modify the MT system as more data become available. These modifications are usually performed in an incremental way \citep{Levenberg10,Denkowski14b,Turchi17}, learning from user post-edits. This incremental learning has also been applied to IMT, either to phrase-based statistical machine translation (SMT) systems \citep{Nepveu04,Ortiz16} or NMT \citep{Peris18}.

The translation of large volumes of data is a scenario very appropriate for the AL framework \citep{Cohn94,Olsson09,Settles09}. The application of AL to SMT has been studied for pool-based \citep{Haffari09,Bloodgood10} and stream-based \citep{Gonzalez11} setups. Later works \citep{Gonzalez12,Gonzalez14}, combined AL together with IMT, showing that AL can effectively reduce the human effort required for achieving a certain translation quality.

All these works were based on SMT systems. However, the recently introduced NMT paradigm \citep{Sutskever14,Bahdanau15} has irrupted as the current state-of-the-art for MT \citep{Bojar17}. Several works aimed at building more productive NMT systems. Related to our work, studies on interactive NMT systems \citep{Knowles16,Peris17a,Hokamp17} proved the efficacy of this framework. A body of work has been done aiming to build adaptive NMT systems, which continuously learn from human corrections \citep{Turchi17,Peris18}. Recently, \citet{Lam18} applied AL techniques to an INMT system, for deciding whether the user should revise a partial hypothesis or not.
However, to our knowledge, a study on the use of AL for NMT in a scenario of translation of unbounded data streams is still missing.
 
	\section{Neural machine translation}
\label{sec:NMT}

NMT is a particular case of sequence-to-sequence learning: given a sequence of words from the source language, the goal is to generate another sequence of words in the target language. This is usually done by means of an encoder--decoder architecture \citep{Sutskever14,Vaswani17}. In this work, we use a recurrent encoder--decoder system with long short-term memory (LSTM) units \citep{Hochreiter97} and an attention mechanism \citep{Bahdanau15}.

Each element from the input sequence is projected into a continuous space by means of an embedding matrix. The sequence of embeddings is then processed by a bidirectional \citep{Schuster97} LSTM network, that concatenates the hidden states from forward and backward layers and produces a sequence of annotations.

The decoder is a conditional LSTM (cLSTM) network \citep{Peris18}. A cLSTM network is composed of several LSTM transition blocks with an attention mechanism in between. We use two LSTM blocks. 

The output of the decoder is combined together with the attended representation of the input sentence and with the word embedding of the word previously generated in a deep output layer~\citep{Pascanu14}. 
Finally, a softmax layer computes a probability distribution over the target language vocabulary.

The model is jointly trained by means of stochastic gradient descent (SGD) \citep{Robbins51}, aiming to minimize the cross-entropy over a bilingual training corpus. SGD is usually applied to mini-batches of data; but it can be also applied sample-to-sample, allowing the training of the NMT system in an incremental way \citep{Turchi17}.

For decoding, the model uses a beam search method \citep{Sutskever14} for obtaining the most probable target sentence $\hat{\y}$, given a source sentence $\x$:
\begin{equation}
\hat{\y}  = \argmax_{\y} p(\y\mid \x) 
\label{eq:search}
\end{equation}

\subsection{Interactive machine translation}
\label{sec:INMT}
As previously discussed, MT systems are not perfect. Their outputs must be corrected by a human agent in a post-editing stage, in order to achieve high-quality translations.

The IMT framework constitutes a more efficient alternative to the regular post-editing. 
In a nutshell, IMT consists in an iterative process in which, at each iteration, the user introduces a correction to the system hypothesis. The system takes into account the correction and provides an alternative hypothesis, considering the feedback from the user. 

In this work, we use a prefix-based IMT protocol: the user corrects the left-most wrong character of the hypothesis. With this action, the user has also validated a correct prefix. Then, the system must complete the provided prefix, generating a suitable suffix. \cref{fig:offline-INMT} shows an example of the prefix-based IMT protocol.

\begin{figure}[!h]
	\centering
	\def\arraystretch{1.4}
	\small
	\begin{tabular}{ccl}
		\toprule
		\multicolumn{2}{l}{\textbf{Source ($\x$)}:}& They are lost forever . \\
		\multicolumn{2}{l}{\textbf{Target ($\hat{\y}$)}:}& Ils sont perdus à jamais . \\
		\hline
		\textbf{IT-0} & \textbf{MT} & Ils sont perdus pour toujours . \\
		\midrule
		\multirow{2}{*}{\textbf{IT-1}} & $\bf{User}$& \textit{\color{darkgreen}{Ils sont perdus }}\regularbox{à}pour toujours .\\
		& $\bf{MT}$ & \textit{\color{darkgreen}{Ils sont perdus à}} jamais . \\
		\midrule
		\textbf{IT-2} & \textbf{User} &\textit{\color{darkgreen}{Ils sont perdus à jamais .}} \\		
		\bottomrule
	\end{tabular}
	
	\caption{\label{fig:offline-INMT} IMT session to translate a sentence from English to French. \textbf{IT-} is the number of iterations of the process. The \textbf{MT} row shows the MT hypothesis in the current iteration. In the \textbf{User} row is the feedback introduced by the user: the corrected character (boxed). We color in green the prefix that the user inherently validated with the character correction. 
	}
\end{figure}

More formally, the expression for computing the most probable suffix ($\hat{\y}_s$) is:
\begin{gather}
\hat{\y}_s = \argmax_{\y_s} p (\y_s \mid \x, \y_p) 
\label{eq:pref}
\end{gather}
where $\y_p$ is the validated prefix provided by the user and $\x$ is the source sentence. Note that this expression is similar to \cref{eq:search}. The difference is that now, the search space is the set of suffixes that complete $\y_p$. 

For NMT systems, \cref{eq:pref} is implemented as a beam search, constrained by the prefix provided by the user \citep{Peris17a,Peris18}.

	\section{Active learning in machine translation}
\label{sec:AL-MT}
When dealing with potentially unbounded datasets, it becomes prohibitively expensive to manually supervise all the translations.
Aiming to address this problem, in the AL framework, a sampling strategy selects a subset of sentences worth to be supervised by the user. Once corrected, the MT system adapts its models with these samples. 

Therefore, the AL protocol applied to unbounded data streams is as follows \citep{Gonzalez12}: first, we retrieve from the data stream ${\cal S}$ a block ${\cal B}$ of consecutive sentences, with the function $\mathtt{getBlockFromStream}({\cal S})$. According to the  $\mathtt{sampling}({\cal B}, \varepsilon)$ function, we select  from ${\cal B}$ a subset ${\cal V}$ of $\varepsilon$ instances, worth to be supervised by the user. See \cref{sec:sampling-strat} for deeper insights on the sampling functions used in this work. These sampled sentences are interactively translated together with the user (\cref{sec:INMT}). This process is done in the function ${\mathtt{INMT}(\theta, \x, \y)}$. Once the user translates via INMT a source sentence $\x$, a correct translation $\hat{\y}$ is obtained. Then, we use the pair $(\x, \hat{\y})$ to retrain the parameters $\theta$ from the NMT model, via SGD. This is done with the function $ \mathtt{update}(\theta, (\x, \hat{\y}))$. Therefore, the NMT system is incrementally adapted with new data. The sentences considered unworthy to be supervised are automatically translated according to according \cref{eq:search}, with the function $\mathtt{translate}(\theta, \x)$. Once we finish the translation of the current block $\cal B$, we start the process again. \cref{alg:AL} details the full procedure.

\begin{algorithm}[!t]
	
	\SetKwInOut{Input}{input}
	\SetKwInOut{Output}{output}
	\SetKwInOut{Aux}{auxiliar}
	\SetKwFunction{ta}{translate}
	\SetKwFunction{sample}{sampling}
	\SetKwFunction{inmt}{INMT}
	\SetKwFunction{update}{update}	
	\SetKwFunction{out}{output}
	\SetKwFunction{get}{getBlockFromStream}
	\SetKwInOut{KwAux}{Auxiliar}
	
	\Input{$\theta$ (NMT model)\\
		${\cal S}$ (stream of source sentences)\\
		$\varepsilon$ (effort level desired) }
	
	\Aux{${\cal B}$ (block of source sentences)\\
		${\cal V}\subseteq {\cal B}$ (sentences to be supervised by the user)
		
	}
	\Begin{
		\Repeat{${\cal S} \neq \emptyset$}{
			
			${\cal B} = \get({\cal S})$\;
			${\cal V} = \sample ({\cal B}, \varepsilon)$\label{alg:sampling-line}\;
			\ForEach{$\x \in {\cal B}$}
			{$\y = \ta(\theta, \x)$\label{alg:translate-line}\;
				\uIf{$\x \in {\cal V}$ }
				{
					$\hat{\y} = \inmt(\theta, \x, \y)$\;
					$\theta= \update(\theta, (\x, \hat{\y}))$\label{alg:retrain-line}\;
					$\out(\hat{\y})$\;
				}
				
				\Else{
					$\out(\y)$\;
				}
			}
		}
	}
	\caption{\label{alg:AL}Active learning for unbounded data streams with interactive neural machine translation.}
\end{algorithm}

\section{Sentence sampling strategies}
\label{sec:sampling-strat}
One of the key elements of AL is to have a meaningful strategy for obtaining the most useful samples to be supervised by the human agent. This requires an evaluation of the informativeness of unlabeled samples. The sampling strategies used in this work belong to two major frameworks: uncertainty sampling~\citep{Lewis94} and query-by-committee~\citep{Seung92}.

As baseline, we use a random sampling strategy: sentences are randomly selected from the data stream ${\cal S}$. Although simple, this strategy usually works well in practice. In the rest of this section, we describe the sampling strategies used in this work.

\subsection{Uncertainty sampling}
The idea behind this family of methods is to select those instances for which the model has the least confidence to be properly translated. Therefore, all techniques compute, for each sample, an uncertainty score. The selected sentences will be those with the highest scores.

\subsubsection*{Quality estimation sampling}

A common and effective way for measuring the uncertainty of a MT system is to use confidence estimation \citep{Gandrabur03,Blatz04,Ueffing07}. The idea is to estimate the quality of a translation according to confidence scores of the words.

More specifically, given a source sentence $\x = x_1, \dotsc, x_J$ and a translation hypothesis $\y = y_1, \dotsc, y_I$, a word confidence score ($C_w$) as computed as \citep{Ueffing05}:

\begin{equation}
C_w(\x, y_i) = \max_{0\leq j\leq J} p(y_i\vert x_j)
\end{equation}
where $p(y_i\vert x_j)$ is the alignment probability of $y_i$ and $x_j$, given by an IBM Model 2 \citep{Brown93}. $x_0$ denotes the empty source word. The choice of the IBM Model 2 is twofold: on the one hand, it is a very fast method, which only requires to query in a dictionary. We are in an interactive framework, therefore speed becomes a crucial requirement. On the other hand, its performance is close to more complex methods \citep{Blatz04,Dyer13}.

Following \citet{Gonzalez12}, the uncertainty score for the quality estimation sampling is defined as:
\begin{equation}
C_{\textrm{qe}}(\x,\y) = 1 - \frac{\vert \{ {y}_i \in \y\vert C_w(\x,{y}_i) > \tau_w\}\vert}{\vert \y\vert}
\end{equation}
where $\tau_w$ is a word confidence threshold, adjusted according to a development corpus. $\vert \cdot \vert$ denotes the size of a sequence or set. 

\subsubsection*{Coverage sampling}

One of the main issues suffered by NMT systems is the lack of coverage: the NMT system may not translate all words from a source sentence. This results in over-translation or under-translation problems \citep{Tu16}.

We propose to use the translation coverage as a measure of the uncertainty suffered by the NMT system when translating a sentence. Therefore, we modify the coverage penalty proposed by \citet{Wu16}, for obtaining a coverage-based uncertainty score:

\begin{equation}
C_{\textrm{cov}}(\x, \y) = \frac{\sum_{j=1}^{\vert\x\vert } \log \big( \min (\sum_{i=1}^{\vert \y\vert}\alpha_{i,j}, 1)\big)}{\vert \x \vert}
\end{equation}
where $\alpha_{i,j}$ is attention probability of the $i$-th target word and the $j$-th source word. 

\subsubsection*{Attention distraction sampling}

When generating a target word, an attentional NMT system should attend on meaningful parts of the source sentence. If the system is translating an uncertain sample, its attention mechanism will be \textit{distracted}. That means, dispersed throughout the source sequence. A sample with a great distraction will feature an attention probability distribution with heavy tails (e.g. a uniform distribution). Therefore, for the attention distraction sampling strategy, the sentences to select will be those with highest attention distraction.

For computing a distraction score, we compute the kurtosis of the weights given by the attention model for each target word ${y}_i$: 
\begin{equation}
\textrm{Kurt}(y_i) = \frac{\frac{1}{\vert \x \vert}\sum_{j=1}^{\vert \x \vert}(\alpha_{i, j} - \frac{1}{\vert \x \vert}) ^4 }{	\big(\frac{1}{\vert \x \vert}\sum_{j=1}^{\vert \x \vert}(\alpha_{i, j} - \frac{1}{\vert \x \vert})^2\big)^2}
\end{equation}
being, as above, $\alpha_{i, j}$ the weight assigned by the attention model to the $j$-th source word when decoding the $i$-th target word. Note that, by construction of the attention model, $ \frac{1}{\vert \x \vert}$ is equivalent to the mean of the attention weights of the word $y_i$.

Since we want to obtain samples with heavy tails, we average the minus kurtosis values for all words in the target sentence, obtaining the attention distraction score $C_{\textrm{ad} }$:
\begin{equation}
C_{\textrm{ad}}(\x, \y) = \frac{\sum_{i=1}^{\vert \y \vert} -\textrm{Kurt}(y_i)}{\vert \y \vert} 
\end{equation}

\subsection{Query-by-committee}
This framework maintains a committee of models, each one able to vote for the sentences to be selected. The query-by-committee (QBC) method selects the samples with the largest disagreement among the members of the committee. The level of disagreement of a sample $\x$ measured according to the vote-entropy function \citep{Dagan95}: 

\begin{equation}
C_{\textrm{qbc}}(\x) = -\frac{ \#V(\x)}{\vert {\cal C} \vert } + \log \frac {\#V(\x)}{\vert {\cal C} \vert} 
\end{equation}
where $\#V(\x)$ is the number of members of the committee that voted $\x$ to be worth to be supervised and $\vert {\cal C} \vert$ is the number of members of the committee. If $\#V(\x)$ is zero, we set the value of $C_{\textrm{qbc}}(\x)$ to $-\infty$.

Our committee was composed by the four uncertainty sampling strategies, namely quality estimation, coverage, attention distraction and random sampling. The inclusion of the latter into the committee can be seen as a way of introducing some noise, aiming to prevent overfitting.

	\section{Experimental framework}
\label{sec:ExperimentalFramework}

In order to assess the effectiveness of AL for INMT, we conducted a similar experimentation than the latter works in AL for IMT \citep{Gonzalez14}: we started from a NMT system trained on a general corpus and followed \cref{alg:AL}. This means that the sampling strategy selected those instances to be supervised by the human agent, who interactively translated them. Next, the NMT system was updated in an incremental way with the selected samples.

Due to the prohibitive cost that an experimentation with real users conveys, in our experiments, the users were simulated. We used the references from our corpus as the sentences the users would like to obtain.

\subsection{Evaluation}

An IMT scenario with AL requires to assess two different criteria: translation quality of the system and human effort spent during the process. 

For evaluating the quality of the translations, we used the BLEU (bilingual evaluation understudy) \citep{Papineni02} score. BLEU computes an average mean of the precision of the $n$-grams (up to order $4$) from the hypothesis that appear in the reference sentence. It also has a brevity penalty for short translations.

For estimating the human effort, we simulated the actions that the human user would perform when using the IMT system. Therefore, at each iteration the user must search in the hypothesis the next error, and position the mouse pointer on it. Once the pointer is positioned, the user would introduce the correct character. These actions correspond to a \textit{mouse-action} and a \textit{keystroke}, respectively.

Therefore, we use a commonly-used metric that accounts for both types of interaction: the keystroke mouse-action ratio (KSMR) \citep{Barrachina09}. It is defined as the number of keystrokes plus the number of mouse-actions required for obtaining the desired sentence, divided by the number of characters of such sentence. We add a final mouse-action, accounting for action of accepting the translation hypothesis. Although keystrokes and mouse-actions are different and require a different amount of effort \citep{Macklovitch05}, KSMR makes an approximation and assumes that both actions require a similar effort.

\subsection{Corpora}

To ensure a fair comparison with the latter works of AL applied to IMT \citep{Gonzalez14}, we used the same datasets: our training data was the Europarl corpus \citep{Koehn05}, with the development set provided at the 2006 workshop on machine translation \citep{Koehn06}. As test set, we used the News Commentary corpus \citep{Callison-Burch07}. This test set is suitable to our problem at hand because i. it contains data from different domains (politics, economics and science), which represent challenging out-of-domain samples, but account for a real-life situation in a translation agency; and ii. it is large enough to properly simulate long-term evolution of unbounded data streams. All data are publicly available. We conducted the experimentation in the Spanish to English language direction. \cref{tab:corpora} shows the main figures of our data.

\begin{table} [h]
	\caption{\label{tab:corpora} Corpora main figures, in terms of number of sentences ($\vert S \vert$), number of running words ($\vert W \vert$) and vocabulary size ($\vert V \vert$). k and M stand for thousands and millions of elements, respectively.}
	\centering
	\small
	\begin{tabular}{lllrrr}
		\toprule
		Corpus & Usage & & $\vert S \vert$ & $\vert W \vert$ & $\vert V \vert$ \\
		\midrule
		\multirow{4}{*}{Europarl} & \multirow{2}{*}{Train}  & En & \multirow{2}{*}{2M} & 46M & 106k \\
		& & Es &  & 48M & 160k \\
		& \multirow{2}{*}{Dev.}  & En & \multirow{2}{*}{2k} & 58k & 6.1k \\
		& & Es &  & 61k & 7.7k \\
				
		\midrule
		
		{\centering News } &  \multirow{2}{*}{Test} & En & \multirow{2}{*}{51k} & 1.2M & 35k \\
		Commentary & & Es &  & 1.5M & 49k \\
	\bottomrule
	\end{tabular}
\end{table}

\subsection{NMT systems and AL setup}

Our NMT system was built using NMT-Keras \citep{Peris18b} and featured a bidirectional LSTM encoder and a decoder with cLSTM units. Following \citet{Britz17}, we set the dimension of the LSTM, embeddings and attention model to 512. We applied batch normalizing transform \citep{Ioffe15} and Gaussian noise during training \citep{Graves11}. The $L_2$ norm of the gradients was clipped to $5$, for avoiding the exploiting gradient effect \citep{Pascanu13}. We applied joint byte pair encoding (BPE) \citep{Sennrich16b} to all corpora. For training the system, we used Adam \citep{Kingma14}, with a learning rate of $0.0002$ and a batch size of $50$. We early-stopped the training according to the BLEU on our development set. For decoding, we used a beam of 6.

We incrementally update the system (\cref {alg:retrain-line} in \cref{alg:AL}), with vanilla SGD, with a learning rate of $0.0005$. We chose this configuration according to an exploration on the validation set. 

The rest of hyperparameters were set according to previous works. The blocks retrieved from the data stream contained $500$ samples (according to \citet{Gonzalez12}, the performance is similar regardless the block size). For the quality estimation method, the IBM Model 2 was obtained with \texttt{fast\_align} \citep{Dyer13} and $\tau_w$ was set to $0.4$ \citep{Gonzalez10a}.


	\section{Results and discussion}
\label{sec:results}

A system with AL involves two main facets to evaluate: the improvement on the quality of the system and the amount of human effort required for achieving such quality. In this section, we compare and study our AL framework for all our sampling strategies: quality estimation sampling (QES), coverage sampling (CovS), attention distraction sampling (ADS), random sampling (RS) and query-by-committee (QBC).

\subsection{Active learning evaluation}

First, we evaluated the effectiveness of the application of AL in the NMT system, in terms of translation quality. \cref{fig:AL-bleu} shows the BLEU of the initial hypotheses proposed by the NMT system (\cref{alg:translate-line} in \cref{alg:AL}), as a function of the percentage of sentences supervised by the user ($\varepsilon$ in \cref{alg:AL}). That means, the percentage of sentences used to adapt the system.  
The BLEU of a static system without AL was $34.6$. Applying AL, we obtained improvements up to $4.1$ points of BLEU.

\pgfplotsset{every tick label/.append style={font=\footnotesize}}

\begin{figure}[h]
		\centering
\footnotesize
	\begin{tikzpicture} 
	\begin{axis}[
	legend columns=5, 
	legend style={
		legend entries =  {QES, ADS, CovS, QBC, RS},
		legend style={/tikz/column 2/.style={column sep=5pt,},draw=none,font=\footnotesize, at={(0.42,1.2)},anchor=north},
	},
	ylabel={\footnotesize BLEU [\%]}, 
	xlabel={\footnotesize Sentences supervised [\%]},
	ymin=34.6,
	ymax=39,
	height=0.25\textheight,	
	width=0.5\textwidth,
	enlarge x limits=false,
	cycle list name=color list
	]	
	
	

	\addplot [smooth, dashdotted, thick, red!75] coordinates {
						  (0,34.6)
					 	  (10,36.3)
					 	  (20,37.0)	
					 	  (30,37.5)				 
					 	  (40,37.8)
					 	  (50,38.0)
					 	  (60,38.2) 
					 	  (70,38.4)
					 	  (80,38.5)
					 	  (90,38.5)
					 	  (100,38.5)			 	  
					 	   };
	
	\addplot [smooth, dashed, thick, blue!80]coordinates {
						  (0,34.6)
						  (10,36.8) 	
						  (20,37.4) 
						  (30,37.6)
						  (40,37.9)
						  (50,38.1)			  
						  (60,38.3)
						  (90, 38.7)						
						  (100,38.7)
						   };   

	\addplot [smooth, densely dotted, thick, darkgreen!80] coordinates {(0,34.6)
                          (10,36.5)
                          (20,37.5)
                          (30,37.7)
                          (40,37.9)
                          (50,38.2)
                          (60,38.4)
                          (70, 38.5)
                          (80,38.5)
                          (90,38.61)
                          (100,38.57)
		  };

\addplot [smooth, densely dotted, thick, orange!80] coordinates {
	(0,34.6) 
	(10,36.9) 
	(20,37.4) 
	(30, 37.678256012)	
	(40, 37.91018)	
	(50, 38.1)			
	(60,38.3) 						      
	(70,38.4) 			
	(90, 38.6)							      
	(100,38.53)
};

\addplot [smooth, densely dashdotted, thick, black!60] coordinates {
	(0,34.6) 
	(10,35.7) 
	(20,37.0) 
	(30, 37.5)	
	(40, 37.8)	
	(50, 38.0)			
	(60,38.2)     
	(70, 38.3)
	(80,38.35)
	(90,38.4)	      
	(100,38.45)
};

	\end{axis} 
\end{tikzpicture}
	\caption{\label{fig:AL-bleu}BLEU of the initial hypotheses proposed by the the NMT system as a function of the amount of data used to adapt it. The percentage of sentences supervised refers to the value of $\varepsilon$ with respect to the block size.
		}
\end{figure}
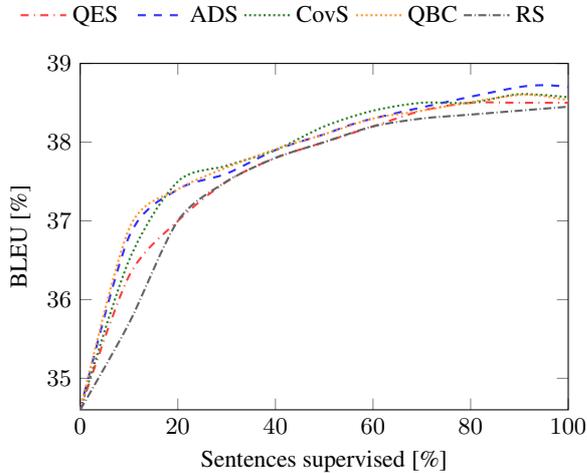

As expected, the addition of the new knowledge had a larger impact when applied to a non-adapted system. Once the system becomes more specialized, a larger amount of data was required to further improve.

The sampling strategies helped the system to learn faster. Taking RS as a baseline, the learning curves of the other techniques were better, especially when using few (up to a $30\%$) data for fine-tuning the system.
The strategies that achieved a fastest adaptation were those involving the attention mechanism (ADS, CovS and QBC). This indicates that the system is learning from the most useful data. The QES and RS required more supervised data for achieving the comparable BLEU results. 
When supervising high percentages of the data, we observed BLEU differences. This is due to the ordering in which the selected sentences were presented to the learner. The sampling strategies performed a sort of curriculum learning \citep{Bengio09}.

\subsection{Introducing the human into the loop}

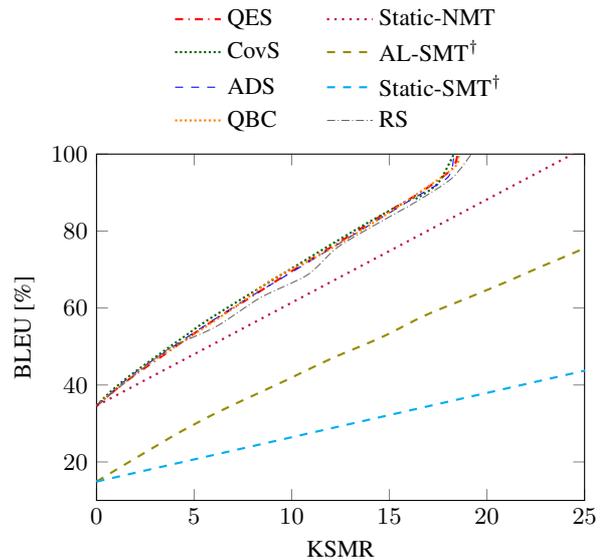
\begin{figure}[!h]
	\centering
\footnotesize
\begin{tikzpicture}[spy using outlines={magnification=2.5}]
\begin{axis}[
legend columns=5, 
transpose legend,
legend style={
legend style={/tikz/column 2/.style={column sep=15pt,},draw=none,   at={(0.5,1.45)},anchor=north},
	},
legend cell align={left},	
ylabel={\footnotesize BLEU [\%]}, 
xlabel={\footnotesize KSMR},
no markers,
xmin=0,
xmax=25,
ymin=10,
ymax=100,
height=0.25\textheight,	
width=0.5\textwidth,
enlarge x limits=false,
cycle list name=color list
]



%


\addplot [ dashdotted, thick,  smooth, red] coordinates {
	(0,34.6)	
	(1.5,40.9)  
	(3.5,48.0)	
	(5.6,55.5)	
	(7.8,62.5)	
	(10.1,69.9)	
	(12.4,77.0)	
	(14.6,84.0)	
	(16.7,90.4)	
	(18.2,96.1)	
	(18.5,100) 	
}; \addlegendentry{QES};

%

\addplot [densely dotted, thick,  smooth, darkgreen] coordinates {
	(0,34.6)	
	(0.7, 38.2) 
	(1.7, 42.4) 
	(3.0, 47.2) 
	(4.5, 52.6) 
	(6.2, 58.5) 
	(13.1, 79.8)
	(16.1, 88.3)
	(16.4, 88.5)
	(17.5, 92.5)
	(18.3,100) 	
	
};\addlegendentry{CovS};

\addplot  [ dashed, smooth,  blue] coordinates {
	(0,34.6)	
	(1.9, 42.9) 
	(3.9, 49.9) 
	(5.9, 56.6)
	(7.9, 63.0)
	(10.0, 69.3)
	(16.0, 87.8)
	(18.0, 94.0) 
	(18.3,100) 	

}; \addlegendentry{ADS};

\addplot  [densely dotted, thick, smooth,  orange]  coordinates {
	(0,34.6)	  
	(2.4, 44.5) 
	(5.1, 53.5) 
	(7.3, 61.0)	
	(9.0, 67.4) 
	(11.3, 73.4) 
	(13.5, 80.0) 
	(15.5, 86.5) 
	(18.3, 96.4) 
	(18.5,100)  

};\addlegendentry{QBC};

\addplot [densely dashdotted,  smooth, black!60] coordinates {
	(0, 34.6)	  
	(2.0, 42.8)  
	(3.8, 49.7)  
	(6.1, 55.1)  
	(8.4, 62.7)  
	(10.8, 68.6)  
	(12.3, 75.6)  
	(14.2, 81.4)  
	(16.4, 87.7) 
	(18.2, 93.7 )
	(19.2, 100)  
};\addlegendentry{RS};

\addplot  [dotted, thick, color=purple] coordinates {(0,34.6) 
	(24.4,100) }; \addlegendentry{Static-NMT};

\addplot  [thick , dashed,  smooth, color=olive]  coordinates {
	(0,14.9)
	(3.8,26.4)
	(6.0,32.4)
	(8.0,37.2)
	(9.7,41.2)
	(12.0,46.8)
	(14.6,52.4)
	(16.8,58.0)
	(19.1,62.8)
	(21.0,66.8)
	(22.7,70.6)
	(24.3,74.0)
	(25.7,77.2)
	(26.9, 80.2)	
	(28.4, 82.8)
	(29.7, 85.8)					  
	(30.8, 88.2)	
	(31.8, 90.0)	
	(32.6, 91.8)	
	(33.5, 93.8)	
	(34.2, 95.2)	
	(35.9, 100)	
}; \addlegendentry{AL-SMT$^\dagger$};
\addplot  [dashed,thick,color=cyan]  coordinates {
	(0,14.9)
	(40, 61)	
}; \addlegendentry{Static-SMT$^\dagger$};

\coordinate (spypoint) at (axis cs:9,64);
\coordinate (magnifyglass) at (axis cs:30,39);
\end{axis} 
\end{tikzpicture}
	\caption{\label{fig:bleu-vs-ksmr}Translation quality (BLEU) as a function of the human effort (KSMR) required. Static-NMT relates to the same NMT system without AL. $^\dagger$ denotes systems from \citet{Gonzalez14}: Static-SMT is a SMT system without AL and AL-SMT is the coverage augmentation SMT system.}
\end{figure}

From the point of view of a user, it is important to assess not only the quality of the MT system, but also the effort spent to obtain such quality. 
\cref{fig:bleu-vs-ksmr} relates both, showing the amount of effort required for obtaining a certain translation quality. We compared the results of system with AL against the same NMT system without AL and with two other SMT systems, with and without AL, from \citet{Gonzalez14}. 

Results in \cref{fig:bleu-vs-ksmr} show consistent positive results of the AL framework. In all cases, AL reduced the human effort required for achieving a certain translation quality. Compared to a static NMT system, approximately a $25\%$ of the human effort can be spent using AL techniques.

Regarding the different sampling strategies, all of them behaviored similarly. They provided consistent and stable improvements, regardless the level of effort desired ($\varepsilon$). This indicates that, although the BLEU of the system may vary (\cref{fig:AL-bleu}), this had small impact on the effort required for correcting the samples. All sampling strategies outperformed the random baseline, which had a more unstable behavior. 
 
Compared to classical SMT systems, NMT performed surprisingly well. Even the NMT system without AL largely outperformed the best AL-SMT system. This is due to several reasons: on the one hand, the initial NMT system was much better than the original SMT system ($34.6$ vs. $14.9$ BLEU points). Part of this large difference were presumably due to the BPE used in NMT: the data stream contained sentences from different domains, but they can be effectively encoded into known sequences via BPE. The SMT system was unable to handle well such unseen sentences. On the other hand, INMT systems usually respond much better to the human feedback than interactive SMT systems \citep{Knowles16,Peris17a}. Therefore, the differences between SMT and NMT were enlarged even more.

Finally, it should be noted that all our sampling strategies can be computed 
speedily. They involve analysis of the NMT attention weights, which are computed as a byproduct of the decoding process; or queries to a dictionary (in the case of QES). The update of NMT system is also fast, taking approximately $0.1$ seconds. This makes AL suitable for a real-time scenario. 

%
%
%
%
%

	\section{Conclusions and future work}
\label{sec:Conclusions}
We studied the application of AL methods to INMT systems. The idea was to supervise the most useful samples from a potentially unbounded data stream, while automatically translating the rest of samples. We developed two novel sampling strategies, able to outperform other well-established methods, such as QES, in terms of translation quality of the final system. 

We evaluated the capabilities and usefulness of the AL framework by simulating real-life scenario, involving the aforementioned large data streams. AL was able to enhance the performance of the NMT system in terms of BLEU. Moreover, we obtained consistent reductions of approximately a $25\%$ of the effort required for reaching a desired translation quality. Finally, it is worth noting that NMT outperformed classical SMT systems by a large margin.

We want to explore several lines of work in a future. First, we intend to apply our method to other datasets, involving linguistically diverse language pairs and low-resource scenarios, in order to observe whether the results obtained in this work hold. We also aim to devise more effective sampling strategies. To take into account the cognitive effort or time required for interactively translating a sentence seem promising objective functions. Moreover, these sampling strategies can be used as a data selection technique. It would be interesting to assess their performance on this task. We also want to study the addition of reinforcement or bandit learning into our framework. Recent works \citep{Nguyen17,Lam18} already showed the usefulness of these learning paradigms, which are orthogonal to our work. Finally, we intend to assess the effectiveness of our proposals with real users in a near future.

	\section*{Acknowledgments}
	The research leading this work received funding from grants PROMETEO/2018/004 and CoMUN-HaT - TIN2015-70924-C2-1-R. We also acknowledge NVIDIA Corporation for the donation of GPUs used in this work.
	
	\bibliographystyle{acl_natbib_nourl}
	\bibliography{main}
	
\end{document}